%% file: main.tex
\documentclass[10pt, conference]{IEEEtran}
\usepackage{amssymb,amsfonts}
\usepackage{graphicx}
\usepackage{textcomp}
\usepackage{xcolor}
\usepackage{cite}
\usepackage[colorinlistoftodos]{todonotes}
\usepackage{booktabs} % For formal tables
\usepackage{bm}
\usepackage{comment}
\usepackage{soul}
\usepackage{float}
\usepackage[normalem]{ulem}
\usepackage[]{algorithm2e}
\usepackage{todonotes}
\usepackage{multirow}
\usepackage{hyperref}
\usepackage{makecell}
\usepackage{tablefootnote}
\usepackage{array}
\usepackage{footmisc}
\usepackage{mdwtab}
\usepackage{algorithm2e}
\usepackage[cmex10]{amsmath}
\DeclareMathOperator*{\argmax}{argmax}
\def\BibTeX{{\rm B\kern-.05em{\sc i\kern-.025em b}\kern-.08em
    T\kern-.1667em\lower.7ex\hbox{E}\kern-.125emX}}

\begin{document}
\title{A Batched Multi-Armed Bandit Approach \\ to News Headline Testing}

\author{\IEEEauthorblockN{Yizhi Mao}
\IEEEauthorblockA{\textit{Oath Inc} \\
Sunnyvale, USA \\
lolam@oath.com}
\and
\IEEEauthorblockN{Miao Chen}
\IEEEauthorblockA{\textit{Oath Inc} \\
Sunnyvale, USA \\
miaoc@oath.com}
\and
\IEEEauthorblockN{Abhinav Wagle}
\IEEEauthorblockA{\textit{Oath Inc} \\
Sunnyvale, USA \\
awagle@oath.com}
\and
\IEEEauthorblockN{Junwei Pan}
\IEEEauthorblockA{\textit{Oath Inc} \\
Sunnyvale, USA \\
jwpan@oath.com}
\and
\IEEEauthorblockN{Michael Natkovich}
\IEEEauthorblockA{\textit{Oath Inc} \\
Sunnyvale, USA \\
mln@oath.com}
\and
\IEEEauthorblockN{Don Matheson}
\IEEEauthorblockA{\textit{Oath Inc} \\
Sunnyvale, USA \\
donm@oath.com}
}

\maketitle

\begin{abstract}
Optimizing news headlines is important for publishers and media sites. A compelling headline will increase readership, user engagement and social shares. At Yahoo Front Page, headline testing is carried out using a test-rollout strategy: we first allocate equal proportion of the traffic to each headline variation for a defined testing period, and then shift all future traffic to the best-performing variation. In this paper, we introduce a multi-armed bandit (MAB) approach with batched Thompson Sampling (bTS) to dynamically test headlines for news articles. This method is able to gradually allocate traffic towards optimal headlines while testing. We evaluate the bTS method based on empirical impressions/clicks data and simulated user responses. The result shows that the bTS method is robust, converges accurately and quickly to the optimal headline, and outperforms the test-rollout strategy by 3.69\% in terms of clicks. 
\end{abstract}


\section{Introduction}

Headline testing is important for publishers and media sites. Visitors to the homepage of online publishers are usually presented with lists or groups of headlines, sometimes along with snippets\footnote{For example, Yahoo Front Page (https://www.yahoo.com/), Google News (https://news.google.com/), The New York Times (https://www.nytimes.com/), and The Wall Street Journal (https://www.wsj.com/).}. A compelling headline would encourage visitors to click it and read the whole article, and thus help increase user engagement, social sharing and revenue. At Yahoo Front Page, we follow a \textit{test-rollout} strategy for headline testing. Once an article is published, multiple title variations are displayed to randomized user buckets with equal size for a defined period to conduct bucket testing. When the test is complete, the headline variant that has most clicks during the bucket testing period is selected and displayed for the rest of the article life. This strategy has been adopted as a common practice to select the best title variants and ad versions for headline testing~\cite{optimizely2014url} and online advertising~\cite{schwartz2017customer}, respectively.

There are two limitations of this test-rollout strategy. First, during the initial testing period, we have to show the title variants with lower click-through rates (CTRs) to a sizable fraction of user population. For those users, we fail to optimize their engagement. Further, a large proportion of article traffic clusters in its early life, because freshness is a key factor of article popularity~\cite{keneshloo2016freshness}. The test-rollout strategy conducts bucket testing at the beginning of the article life, which may lead to significant click loss.

On the other hand, the performance of headline variations usually varies over time. The conclusion drawn from the initial testing period may not always hold throughout the whole article life. The test-rollout practice is not able to capture any changes subsequent to the bucket testing period. 

To address the limitations mentioned above, we formulate headline testing as a multi-armed bandit (MAB) problem, and introduce a batched Thompson Sampling method to optimize user engagement while learning the performance of each headline variant. The MAB problem is defined as follows. There are K arms, each associated with an unknown reward distribution. The player iteratively plays one arm, observe the associated reward, and decides which arm to play in the next iteration~\cite{agrawal2012analysis}. There is a tension between selecting the current best-performing arm to harvest immediate gain (exploitation) and discovering the optimal arm, \textit{i.e.,} the arm with the highest expected reward, but risking immediate loss (exploration). In the headline testing scenario, the headline variants of an article correspond to the arms, and the click count of each headline variant is the reward associated with each arm. Our goal is to maximize the sum of clicks across arms for each article.

There are many MAB algorithms such as $\epsilon$-greedy~\cite{kuleshov2014MABalgorithms}, Upper Confidence Bound (UCB) algorithms~\cite{auer2002UCB1}~\cite{audibert2009UCBV}~\cite{audibert2010MOSS}, Thompson Sampling~\cite{thompson1933likelihood}, and Gittins Index~\cite{gittins2011GI}. Among them we select Thompson Sampling due to its strong empirical results~\cite{chapelle2011empirical}~\cite{scott2010modern}, solid theoretical guarantees~\cite{granmo2008two-armed}~\cite{may2012optimistic}~\cite{agrawal2012analysis}~\cite{kaufmann2012thompson}, and wide industrial application~\cite{graepel2010microsoft}~\cite{agarwal2013linkedinway}~\cite{agarwal2014laser}~\cite{scott2015multi}~\cite{hill2017amazon}.

In traditional Thompson Sampling, model parameters are updated for every single user response. It becomes a formidable computational burden, as our site has a high volume of incoming traffic at high velocity. This leads us to consider an algorithm that processes user responses after they arrive in batches over a certain time period. Thompson Sampling with batch updates had been studied in display advertising and news article recommendation to analyze how it performs in the case of delayed user response processing compared with other MAB algorithms~\cite{chapelle2011empirical}. Batch update is incorporated in recent industrial applications of Thompson Sampling~\cite{hill2017amazon}~\cite{scott2014url}. Although how they select the update frequency is not disclosed, \cite{hill2017amazon} adopts a method to update traffic allocation once a day, and~\cite{scott2014url} updates twice a day. 

In this paper, we present a \textit{batched Thompson Sampling (bTS)} method that is tuned for optimal performance when user feedback (\textit{i.e.,} observed rewards) is processed in batches. The performance evaluation is based on empirical impressions/clicks of articles at Yahoo Front Page and user responses simulated from empirical CTRs, and shows that the bTS method is robust, converges quickly to the true optimal arms, and outperforms the test-rollout strategy. Our study is motivated by the headline testing problem, but one can apply this algorithm to other real-world problems where we need to accomplish optimization while testing with high volume and high velocity incoming data.

The rest of the paper is organized as follows. Section~\ref{currentpractice} introduces the current headline testing practice at Yahoo Front Page, as well as its limitations. Section~\ref{methodology} describes the batched Thompson Sampling (bTS) method that we propose to apply in headline testing to gain more reward. We show the evaluation of the bTS method in Section~\ref{evaluation}. Section~\ref{futurework} concludes and discusses the future work.  

\section{Current Headline Testing Practice} \label{currentpractice}
\input{2.currentpractice.tex}

\section{Methodology} \label{methodology}
\input{3.methodology.tex}

\section{Methodology Evaluation} \label{evaluation}
\input{4.evaluation.tex}

\section{Conclusions and Future Work} \label{futurework}
\input{5.futurework.tex}

\section*{Acknowledgment} \label{acknowledgment}
\input{6.acknowledgment.tex}

\bibliographystyle{IEEEtran}

\bibliography{IEEEabrv,main.bib}

\end{document}

%% file: 2.currentpractice.tex
Currently at Yahoo Front Page, headline testing follows the test-rollout strategy which consists of two periods: 

\begin{itemize}
\item \textit{Testing period} is the first hour after an article is published. During the testing period, each viewing request is randomly assigned to one of the headline variants with equal probability. At the end of the testing period, we deem the headline variant with the highest CTR as the \textit{winner headline}.
\item \textit{Post-testing period} refers to the remaining article lifespan after the testing period, during which we display the \textit{winner headline} to all traffic.
\end{itemize}

The test-rollout strategy for headline testing is intuitive and easy to implement in the system. However, it has the following two limitations:

\subsection{User Engagement Loss in the Testing Period}
\label{limit1-fixedtraffic}

In the first hour testing period, impressions are equally allocated to each headline variant, which means that for an article with $K$ arms, only $\frac{1}{K}$ of the traffic is assigned to the headline variant with the highest underlying CTR. Showing inferior headline variants to $\frac{K-1}{K}$ of the traffic will sacrifice user engagement (\textit{e.g.,} article clicks) and user experience. The loss of user engagement and experience are both sizable because the traffic in the testing period covers as high as 24.36\% of the total impressions across all articles. This result is based on empirical headline testing data from Yahoo Front Page, which will be described in Section~\ref{empdata}.

\subsection{Arm Performance Discrepancy Between Testing and Post-testing Period}
\label{limit2-fluctuateCTR}
The test-rollout practice is unable to capture any performance change of headline variations beyond the testing period. Empirical data from Yahoo Front Page, same as the data mentioned above, shows the headlines deemed best during the testing period changed their performance afterward: on average there is a 12\% discrepancy in their CTR between testing and post-testing periods. Thus, it is possible for the arm deemed optimal during the one-hour testing period to be actually sub-optimal over the article lifespan. Such a performance change cannot be observed and handled in the current practice. 

To enhance this test-rollout practice, we formulate headline testing as an MAB problem and present a batched Thompson Sampling approach. It is able to explore for the optimal variant, and at the same time gradually shift traffic to the best-performing variant. The following section introduces the MAB headline testing methodology in detail.

%% file: 3.methodology.tex
This section describes the \textit{batched Thompson Sampling (bTS)} method that is able to gradually allocate traffic towards the well-performing arms, while leaving some traffic to other arms so as to explore for the possibly unobserved optimal arm. Section~\ref{preliminaries} formulates headline testing as a Bernoulli bandit problem, and introduces Thompson Sampling for Bernoulli bandits. In Section~\ref{batchupdates}, we explain the rationale of incorporating batch updates in Thompson Sampling, and introduce the factors associated with bTS. Section~\ref{batchupdates} also describes our empirical impressions/clicks data and how we simulate user responses based on them. Sections~\ref{factor1} to~\ref{factor3} present how we determine and tune the factors of bTS.

\subsection{Preliminaries}
\label{preliminaries}
\subsubsection{Headline Testing as a Bernoulli Bandit Problem}
\label{problemformulation}
Suppose an article has $K$ headline variants written by editors. In the MAB framework, each headline variant is treated as an arm. Each headline, when displayed, yields either a click (success) or no click (failure) as the reward. The reward for headline $k \in \{ 1, ..., K\}$ is Bernoulli distributed, with the \textit{success probability} (\textit{i.e.,} the probability of being clicked) as $\theta_k$ $\in [0,1]$. The reward distribution of arm $k$ is fixed but unknown, in the sense that its parameter $\theta_k$ is unknown. At time step $t \in [1,T]$, we select an arm to display, and collect the reward observed from the selected arm. Here $T$ is the total number of impressions we decide to run MAB experiment on. Our goal is to maximize the total clicks for this article.

\subsubsection{Thompson Sampling}
\label{TSintro}
Thompson Sampling is a randomized Bayesian algorithm to solve the MAB problem of reward maximization~\cite{thompson1933likelihood}. The general idea of Thompson Sampling is to impose a prior distribution on the parameters of the reward distribution, update the posterior distribution using the observed reward, and play an arm according to its posterior probability at each time step. 

In Bernoulli bandit problems, Thompson Sampling uses Beta distribution to model the success probability $\theta_k$ for arm $k$ because the observed reward follows a Bernoulli distribution, and the Beta distribution is a conjugate prior for the Bernoulli distribution~\cite{agrawal2012analysis}. Initially, the Thompson Sampling algorithm imposes a Beta(1,1) prior on the success probabilities of all arms. It is a reasonable initial prior, because Beta(1,1) is the uniform distribution on the interval (0,1)~\cite{agrawal2012analysis}~\cite{chapelle2011empirical}. At time step $t \in [1,T]$, Thompson Sampling algorithm draws a random sample from the Beta distribution of each arm, and displays the arm associated with the largest sampled value. Based on the observed feedback of the displayed arm, the Beta($\alpha^t$, $\beta^t$) distribution of the displayed arm is updated to Beta($\alpha^t$ + 1, $\beta^t$) if the feedback is a click, or to Beta($\alpha^t$, $\beta^t$ + 1) otherwise.

Many studies have demonstrated the strong performance of Thompson Sampling algorithm in the MAB problem, both theoretically and empirically. \cite{granmo2008two-armed} investigates Thompson Sampling as Bayesian Learning Automaton, and shows that in the two-armed Bernoulli bandit problem, Thompson Sampling converges to only playing the optimal arm with probability one. 
\cite{may2012optimistic} proposes an optimistic version of Thompson Sampling, and proves both Thompson sampling and the optimistic version result in optimal behaviour in the long term consistency sense described by~\cite{yang2002randomized}. 
Further, \cite{agrawal2012analysis} and~\cite{kaufmann2012thompson} provide regret bounds for Thompson Sampling that are asymptotically optimal in the sense defined by~\cite{lai1985asymptotically}, so that it has the theoretical guarantee competitive to UCB algorithms. Empirically, \cite{chapelle2011empirical} shows the performance of Thompson Sampling is competitive to or better than that of other alternative MAB algorithms, such as $\epsilon$-greedy and UCB, on real-world problems like display advertising and news article recommendation. \cite{chapelle2011empirical} also mentions Thompson Sampling can be implemented efficiently, in comparison with full Bayesian methods such as Gittins index. Recently, adaptations of Thompson Sampling have been applied in many domains, such as revenue management~\cite{ferreira2017revmmgt}, recommendation system~\cite{kawale2015recommendation}, online service experiments~\cite{scott2015multi}, website optimization~\cite{hill2017amazon}, and online advertising~\cite{graepel2010microsoft}~\cite{agarwal2013linkedinway}~\cite{agarwal2014laser}. 

\subsection{Batch Updates for Real-world High-volume Traffic}
\label{batchupdates}
In traditional Thompson Sampling for Bernoulli bandits, the Beta distribution of the selected arm is updated after every reward feedback is observed. In a real-world system, especially when both the volume and velocity of incoming traffic are high, the feedback is typically processed in batches over a certain period of time. This is the case for our current infrastructure of headline testing. Thus, to implement Thompson Sampling, it is necessary to apply the batched Thompson Sampling (bTS) that updates the posterior distribution after a fixed time period. 

The general procedure of bTS is described as follows: within each fixed time interval (\textit{i.e.,} batch), the Beta distribution of each arm remains unchanged. We allocate traffic across arms based on their random Beta distribution samples drawn for each incoming view event. At the end of a time interval, we aggregate the data collected within this batch, namely the numbers of clicks and impressions for each arm, and use the aggregated data to update the Beta distribution of each arm.

There are three factors to be tuned for bTS. The first one is how long the algorithm should run. We determine an algorithm stopping point after which the click gain is so little that the system cost is not worthwhile. The second is how we aggregate the feedback within each batch to update the posterior distributions. The last factor is the time interval between each update. These three factors are determined based on the empirical as well as simulated data from our real-world headline testing platform, which will be described as follows.

\subsubsection{Empirical data} \label{empdata}
The empirical data cover the articles with headline testing at Yahoo Front Page on two weekdays and one weekend day. In the testing period, the data consist of impressions and clicks of all headline variants for each article. While in the post-testing period, we only have data on the headline variant deemed best in the previous testing period for each article, because all traffic is shifted to this headline variant. The impressions and clicks are aggregated by every minute. 

\subsubsection{Simulation practice}
\label{simulationpractice}

Similar as how~\cite{chapelle2011empirical} evaluates Thompson Sampling in display advertising, we evaluate the performance of bTS under different factor values in a simulated environment, where impressions and CTRs are real, but the user responses are simulated based on the empirical CTR of each headline. 

In the simulation of Thompson Sampling~\cite{chapelle2011empirical}, the reward probability of each arm is modeled by a Beta distribution which is updated after an arm is selected. For batched Thompson Sampling, the Beta distribution of each arm is only updated at the end of each batch, \textit{i.e.}, a fixed-length time interval. Accordingly, the number of events within a batch, which is referred to as \textit{batch size}, is determined by the count of empirical impressions occurred in this fixed-length time interval. The batches of an article rarely have the same size, because the number of impressions usually varies a lot in different time intervals. Table~\ref{table:batchsize} illustrates an example of how batch sizes are calculated based on the minute-level impressions when updates occur every 3 minutes.

\begin{table}
\renewcommand{\arraystretch}{1.3}
\caption{Impressions and batch sizes for a sample article}
\label{table:batchsize}
\centering
	\begin{tabular}{| c | c | c | c |}
	\hline
    Timestamp & Impressions\footnotemark & Batch Index & Batch size \\ 
    \hline
    \hline
    12:48:00	 & 615 & 1 & \multirow{3}{*}{9,945} \\ 
    \cline{1-3}
	12:49:00	 & 4,568 & 1 & \multirow{3}{*}{}\\
    \cline{1-3}
	12:50:00	 & 4,762 & 1 & \multirow{3}{*}{} \\
    \hline
	12:51:00	 & 5,282 & 2 & \multirow{3}{*}{16,028} \\
    \cline{1-3}
	12:52:00	 & 5,412 & 2 & \multirow{3}{*}{} \\ 
    \cline{1-3}
	12:53:00	 & 5,334 & 2 & \multirow{3}{*}{} \\
    \hline
	\end{tabular}
	
\end{table}
\footnotetext{The numbers are for illustration purpose only. They are not actual impressions.} 

We simulate user clicks following the practice of~\cite{schwartz2017customer}:
Upon a viewing request, suppose the algorithm selects to display arm $k$, then the user response  is simulated from the Bernoulli($\hat{\theta}_k$) distribution, where we denote $\hat{\theta}_k$ as the estimated success probability of arm $k$, defined by its empirical CTR during the \textit{testing period}. Note that for the arm deemed best during the testing period, we still use its testing period empirical CTR to calculate $\hat{\theta}_k$, although theoretically we can calculate its ``overall'' empirical CTR including both testing and post-testing period. This is because this ``overall'' empirical CTR is not comparable to the empirical CTRs of other arms, which can only be calculated during the testing period. After the simulation is completed on all articles, we quantify the performance of a headline testing algorithm by total clicks summed across all articles over their lifespans.

\subsection{Factor 1: Algorithm Stopping Point}
\label{factor1}
Due to the observed CTR discrepancy between the testing and post-testing period illustrated in Section~\ref{limit2-fluctuateCTR}, we would avoid stopping the algorithm too early, so as to cover any potential performance change among headline variations. One straightforward proposal is to run the algorithm throughout the whole article life. Although technically achievable, this proposal is not desired, given that the distribution of impressions over time usually has a very long right tail. When the impressions are very sparse, the click gain is so little that it is not worth the engineering overhead of running the algorithm. Thus, we would like to stop the algorithm when the majority of articles are no longer active.

\begin{figure}
\centering
    \includegraphics[width=\linewidth]{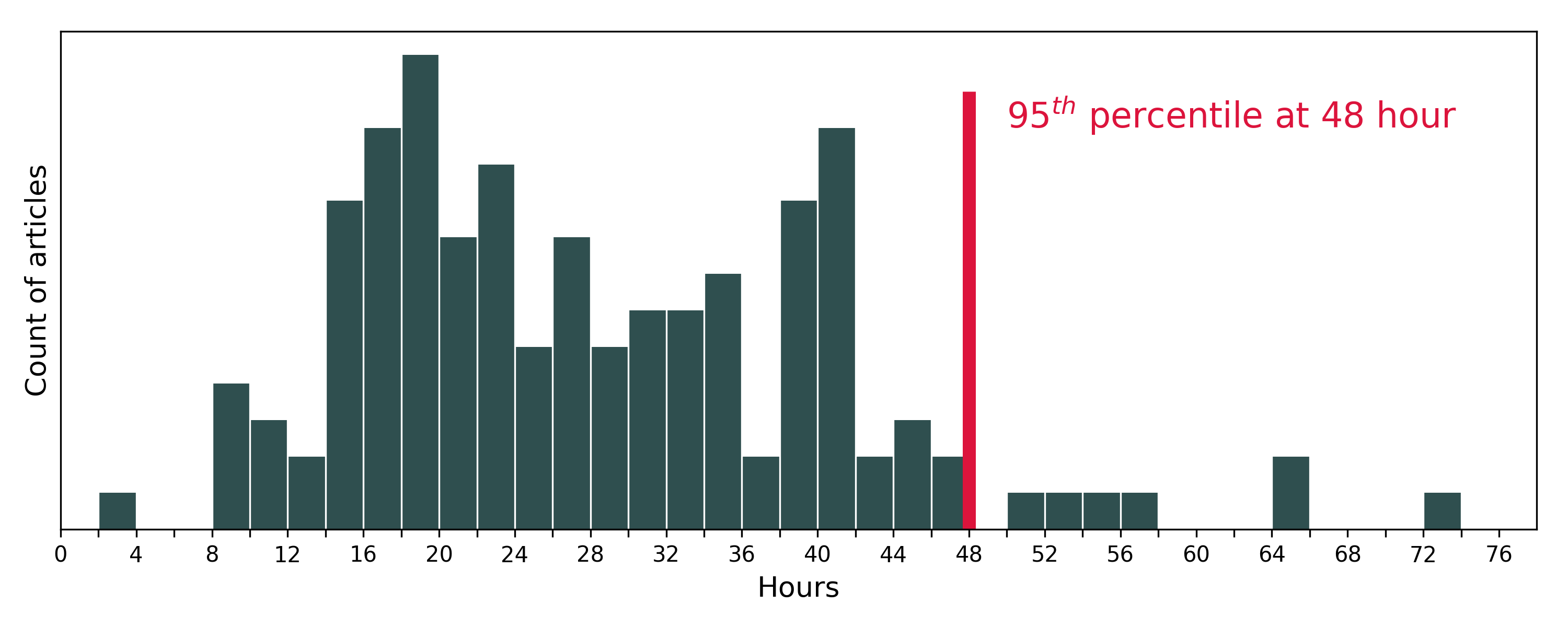}
    \caption{Histogram of article active lifespan}
    \label{fig:activelifespanhist}
\end{figure}

We define the \textit{active lifespan}  of an article as the time it takes to reach 95\% of its total impressions. Fig.~\ref{fig:activelifespanhist} demonstrates that the active lifespans of 95\% of the articles are under 48 hours. Thus, we take 48 hours as the algorithm stopping point. 

\subsection{Factor 2: Update Methods}
\label{factor2}

In traditional Thompson Sampling with Bernoulli bandits, the Beta($\alpha$, $\beta$) distribution of the selected arm is updated to Beta($\alpha + 1$, $\beta$) if we observe a click, otherwise it is updated to Beta($\alpha$, $\beta + 1$). When observed responses come in batches, we need to aggregate the impressions and clicks data for each arm within each batch, before updating the corresponding Beta distribution. We consider two update methods to achieve this: \textit{summation update} and \textit{normalization update}. 

To specifically describe the two update methods, we denote the Beta distribution of arm $k$ in the $t$-th batch by Beta($\alpha^t_{k}$, $\beta^t_{k}$). $S^t_{k}$ and $F^t_{k}$ are the click and non-click counters for arm $k$ in batch $t$. $M^t$ denotes the number of impressions in the $t$-th batch. 
\begin{algorithm} 
\SetAlgoLined

Initialize $\alpha^1_{k} = 1$ and $\beta^1_{k} = 1$ $\forall k$ \;
 \For{batch index $t \in \{1,...,T\}$}{
 Initialize $S^t_{k} = 0$ and $F^t_{k} = 0$ $\forall k$ \;
  \For{event $i \in $ \{1,...,$M^t$\} }{
  	\For{arm $k \in \{1,...,K\}$}{
    	Draw random sample $x_k$ from Beta($\alpha^t_{k}$, $\beta^t_{k}$)}
    Display arm $k^* = \argmax_{k} x_k$ and observe $r$\;
    \eIf{$r=1$}{
   $S^t_{k^*} = S^t_{k^*} + 1$
   }{
   $F^t_{k^*} = F^t_{k^*} + 1$
  }
  }
  \For{arm $k \in \{1,...,K\}$}{
  $\alpha^{t+1}_k$ = $\alpha^t_{k} + S^t_{k}$ \;
  $\beta^{t+1}_k$ = $\beta^t_{k} + F^t_{k}$ \;
 }
  
 }
 \caption{Batched Thompson Sampling with summation update}
 \label{alg:summation}
\end{algorithm}

Algorithm~\ref{alg:summation} explains the bTS method with \textit{summation update}. It is a direct extension from the event-level update method of traditional Thompson Sampling, where $\alpha^t_{k}$ and $\beta^t_{k}$ are updated by raw counts of clicks and non-clicks. 

We also consider another update method named as \textit{normalization update}. As illustrated in Algorithm~\ref{alg:normalization}, it increments $\alpha$ and $\beta$ by the number of normalized clicks and non-clicks respectively, assuming equal traffic allocation across arms. 

\begin{algorithm} 
\SetAlgoLined
Initialize $\alpha^1_{k} = 1$ and $\beta^1_{k} = 1$ $\forall k$ \;
 \For{batch index $t \in \{1,...,T\}$}{
 Initialize $S^t_{k} = 0$ and $F^t_{k} = 0$ $\forall k$ \;
  \For{event $i \in $ \{1,...,$M^t$\}}{
  	\For{arm $k \in \{1,...,K\}$}{
    	Draw random sample $x_k$ from Beta($\alpha^t_{k}$, $\beta^t_{k}$)}
    Display arm $k^* = \argmax_{k} x_k$ and observe $r$\;
    \eIf{$r=1$}{
   $S^t_{k^*} = S^t_{k^*} + 1$
   }{
   $F^t_{k^*} = F^t_{k^*} + 1$
  }
  }
  \For{arm $k \in \{1,...,K\}$}{
  $\alpha^{t+1}_k$ = $\alpha^t_{k} + \frac{M^t}{K} \frac{S^t_{k}}{S^t_{k} + F^t_{k}}$\;
  $\beta^{t+1}_k$ = $\beta^t_{k} + \frac{M^t}{K} (1-\frac{S^t_{k}}{S^t_{k} + F^t_{k}})$ \;
 }
  
 }
 \caption{Batched Thompson Sampling with normalization update}
 \label{alg:normalization}
\end{algorithm}

Normalization update method addresses a side-effect of imbalanced traffic allocation across arms within each batch, which may fail to update in favor of the true optimal arm. More specifically, if arm $k$ has few clicks in a batch, it may be a reflection of a low $\theta_k$, or because the traffic allocated to this arm is not large enough to generate many clicks. Normalization update helps to mitigate the second possibility by assuming each arm has equal traffic allocation. The downside of this method is that it lowers the noise tolerance of the algorithm, because it would give $\frac{1}{K}$ of the weight to potential noise in the data. Given that K is usually less than four, this method may magnify the potential noise. 

We compare the performance of the two methods using the simulation practice described in Section~\ref{simulationpractice}. The algorithm runs for 48 hours as determined in Section~\ref{factor1}, and the performance of either update method is quantified by the total click counts for all articles over their lifespans. 

\begin{table*}
\renewcommand{\arraystretch}{1.3}
\caption{Comparison between summation update and normalization update}
\label{table:updatemethod}
\centering
	\begin{tabular}{| c | c | c | c | c | c | c |} 
	\hline
    \multirow{2}{*}{\makecell{Total Clicks}} & \multicolumn{6}{ c|}{Update frequency in minutes} \\
    \cline{2-7}
    \multirow{2}{*}{} & 1-min &  3-min  & 5-min & 10-min & 30-min & 60-min \\ 
    \hline
    \hline
    \makecell{Summation Update Method\\ versus \\ Normalization Update Method} & +2.50\% & +3.57\% & +1.43\% & +1.08\% & +1.28\% & +0.41\% \\ 
    
    \hline
	\end{tabular}
\end{table*}

According to the result shown in Table~\ref{table:updatemethod}, summation update consistently has better performance than normalization update across different update frequencies. One explanation is that, in the case of headline testing, data noise has more impact on the algorithm performance than the side-effect of imbalanced traffic allocation across arms. Thus, we use the summation update method described in Algorithm~\ref{alg:summation} for the bTS method.

\subsection{Factor 3: Update Frequency}
\label{factor3}

\begin{table*}
\renewcommand{\arraystretch}{1.3}
\caption{Comparison between different update frequencies for summation update method}
\label{table:updatefreq}
\centering
	\begin{tabular}{| c | c | c | c | c | c | c |} 
	\hline
    \multirow{2}{*}{\makecell{Total Clicks}} & \multicolumn{6}{ c|}{Update frequency in minutes} \\
    \cline{2-7}
    \multirow{2}{*}{} & 1-min &  3-min  & 5-min & 10-min & 30-min & 60-min\\ 
    \hline
    \hline
    \makecell{ \% Gap from the\\ Best-performing \\ Update Frequency} & 0 & -0.04\% & -0.06\% & -0.18\% & -0.76\% & -1.52\% \\ 
    \hline
	\end{tabular}
\end{table*}

In Table~\ref{table:updatefreq}, we demonstrate the percentage gap of total clicks between the best-performing update frequency and the remaining frequencies. The table shows that for the bTS method with summation update, more frequent updates lead to more gain in clicks. The gain becomes marginal when the update frequency is lower than 5 minutes. Our infrastructure is capable of updating as frequent as every 5 minutes with almost no cost. However, going beyond 5-minute frequency creates a formidable challenge towards the infrastructure, due to the network transformation cost among multiple components of the system. In consideration of the marginal benefit associated with more granular updates, choosing 5 minutes as the update frequency is a reasonable decision for our use-case.

To conclude the factors we have selected for the bTS method in our real-world headline testing scenario to achieve the trade-off between gain in clicks and system cost: we run the algorithm for 48 hours, aggregate data in each batch using summation update method described in Algorithm~\ref{alg:summation}, and update the Beta distribution of each arm every 5 minutes.

%% file: 4.evaluation.tex
This section evaluates the bTS method in the setting of headline testing. In Sections~\ref{sec:falseconvrate}, \ref{sec:optimizationspeed}, and~\ref{sec:selfcorrection}, we evaluate its false convergence rate, speed of optimization, and robustness, respectively. Then, we show bTS outperforms the test-rollout practice, in terms of click gain in Section~\ref{sec:clickgain}, and less exposure of sub-optimal headlines in Section~\ref{sec:fewer}.

\subsection{False Convergence Rate}
\label{sec:falseconvrate}

We analyze the false convergence rate of the algorithm, defined as the proportion of articles that fail to allocate most traffic to the \textit{optimal arm} -- the arm with highest $\theta$ -- when their traffic allocation across arms is stable. Our evaluation shows 99.25\% of the articles converge correctly for the bTS method.

\begin{figure}
    \centering
    \includegraphics[width=\linewidth]{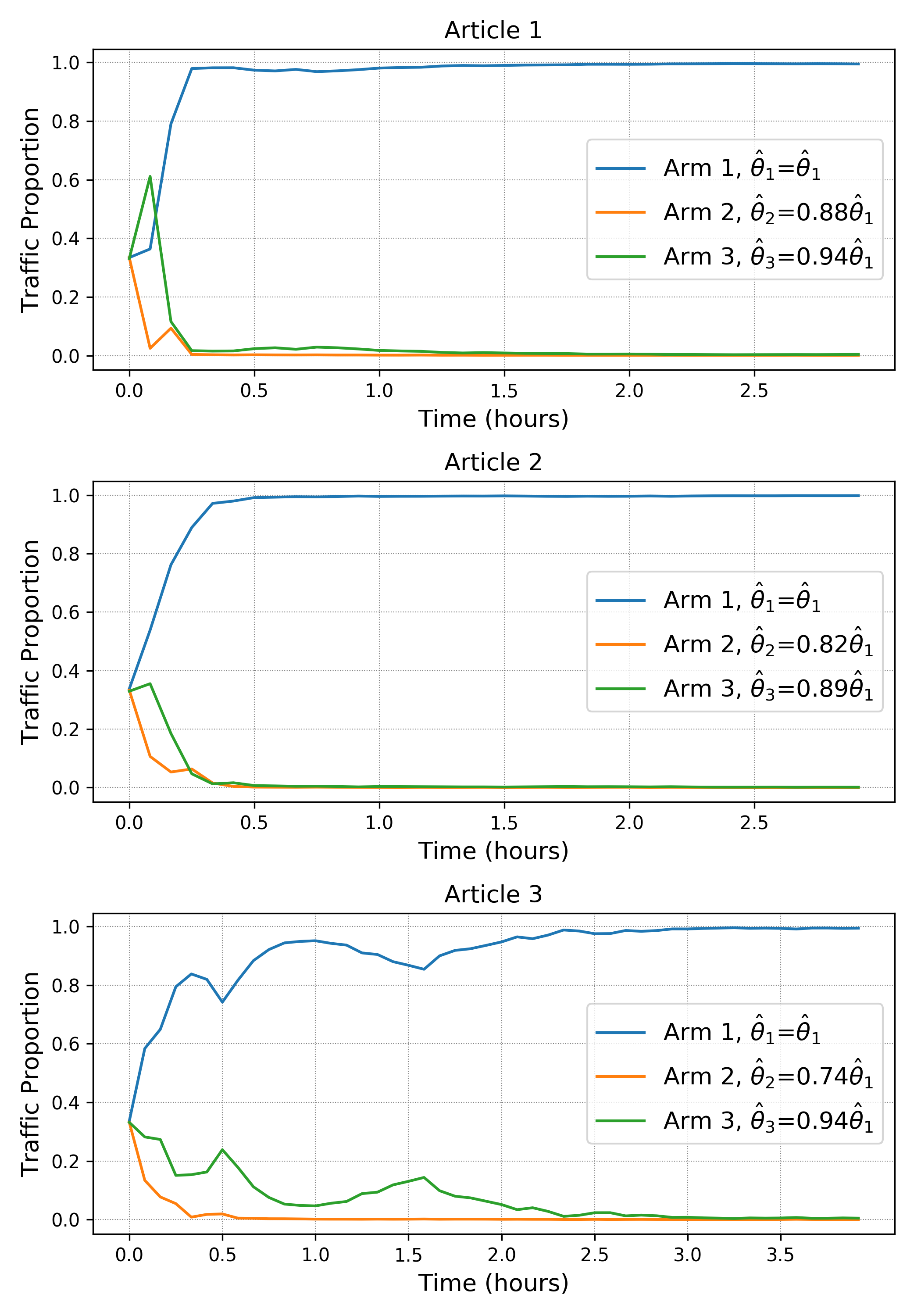}
    \caption{Sample articles with correct convergence: traffic proportion allocated to each arm over time. 
    For each article, the $\hat{\theta}$'s of arms are illustrated as the ratio to the highest $\hat{\theta}$. 
    } 
    \label{fig:conv_right}
\end{figure}

Fig.~\ref{fig:conv_right} demonstrates the traffic proportion over time of some sample articles that converge correctly: initially all arms have equal traffic allocation. When the algorithm begins running, it starts to explore for the optimal arm, while simultaneously allocating more traffic to the arm that performs well. 

Fig.~\ref{fig:conv_right} also illustrates how Thompson Sampling gracefully handles exploration at the level of individual arms, as pointed out in~\cite{scott2015multi}. bTS explores the clearly inferior arms (\textit{e.g.,} Arm 2 in all sample articles) less frequently than arms that may be optimal, \textit{i.e.,} good arms (\textit{e.g.,} Arm 1 and Arm 3 in the sample articles). This increases the click gain by shifting traffic from a clearly inferior arm to arms with better performance. It also helps distinguish the optimal arm faster, because there are more samples to compare among the good arms.

\subsection{Speed of optimization}
\label{sec:optimizationspeed}
\begin{figure}[!t]
\centering
    \includegraphics[width=\linewidth]{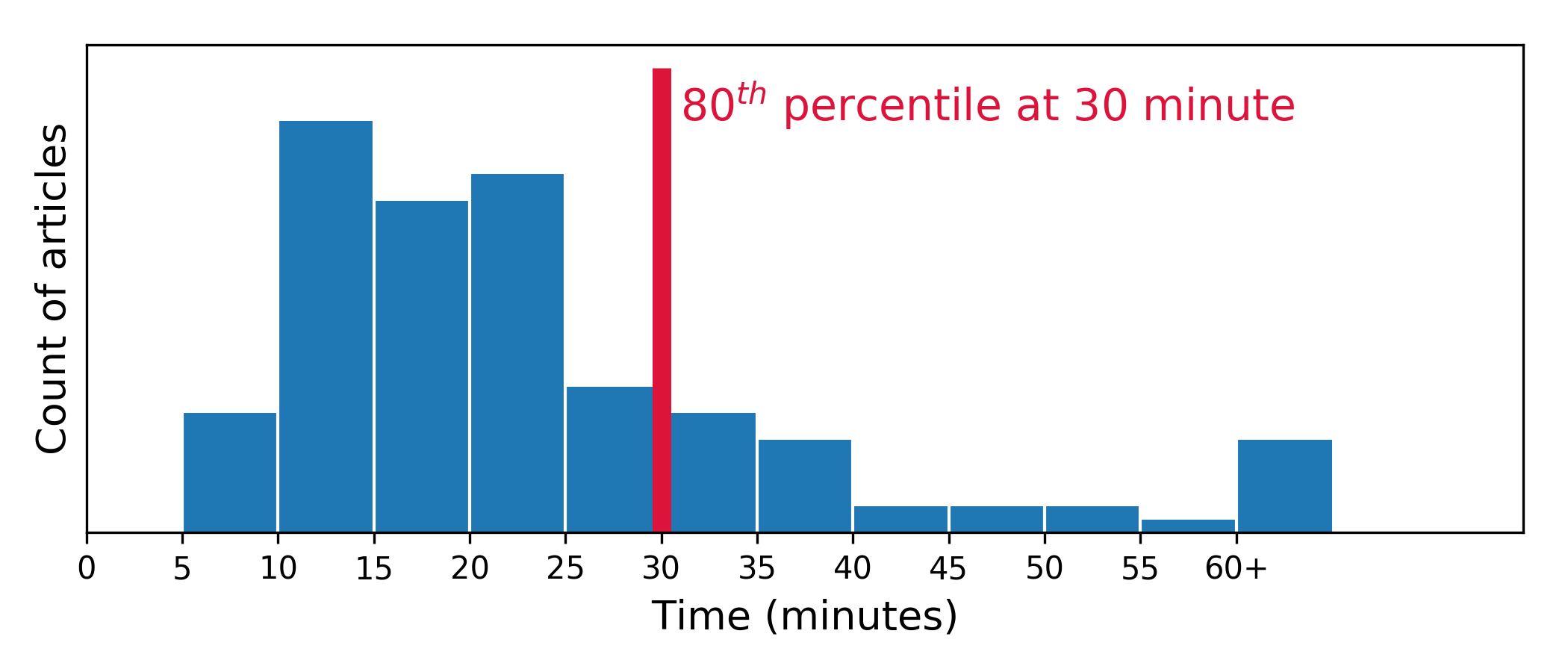}
    \caption{Histogram of time to optimize. All articles that takes over 60 minutes to optimize are in the last bin.}
    \label{hist:optimizationspeed}
\end{figure}

We show the bTS method optimizes quickly, by analyzing the time it takes for the optimal arm to constantly have the largest traffic allocation. The histogram presented in Fig.~\ref{hist:optimizationspeed} shows the distribution of the time to optimize among articles that converge correctly. The $80^{th}$ percentile of the time to optimize is 30 minutes, which means after 30 minutes, 80\% of the articles constantly have the largest traffic proportion allocated to their optimal arms. 

\begin{figure}[!t]
\centering
    \includegraphics[width=\linewidth]{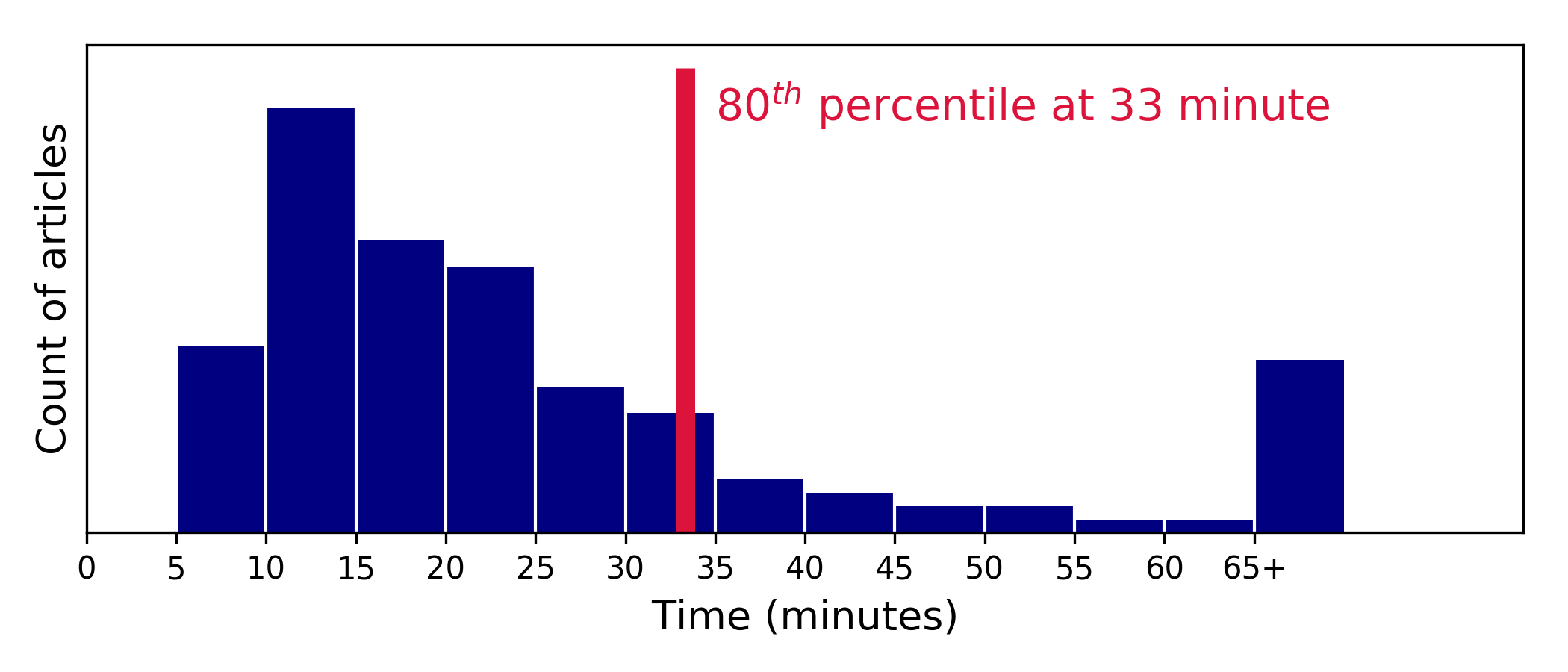}
    \caption{Histogram of time needed for self-correction. All articles that takes over 65 minutes to self-correct are in the last bin.}
    \label{hist:selfcorrection}
\end{figure}

\subsection{Self-correction under Stress Test}
\label{sec:selfcorrection}
We conduct a stress test via simulation to evaluate the robustness of our algorithm against unfavorable traffic allocation. In the stress test, we change the initialized beta distribution for each arm, so that around 90\% of the traffic is allocated to the arm with the lowest $\theta$, and the remaining traffic is equally allocated to the other arms. We analyze the time it takes for the optimal arm to have the highest likelihood to be displayed for five consecutive batches\footnote{This is equivalent to the optimal arm having the highest mean of beta distribution among other arms for five consecutive batches.}, which we refer to as \textit{self-correction}, and illustrate the distribution of the time needed for self-correction in Fig.~\ref{hist:selfcorrection}. The vertical red line shows that the $80^{th}$ percentile is at 33 minutes -- 80\% of the articles are able to self-correct within 33 minutes.

\subsection{Gain in Clicks}
\label{sec:clickgain}
In this section, we illustrate that applying bTS to headline testing generates more clicks than the test-rollout strategy, especially during the testing period of an article when bTS dynamically allocate more traffic to the well-performing arms, while impressions are equally allocated to each arm in the test-rollout strategy. 
The \textit{test-rollout baseline} is set up as follows: for a given article, denote the total baseline click number as $C^{baseline}$, the click number for arm $k$ during the testing period as $C_k^{testing}$, and its click number during the post-testing period as $C_k^{post}$, then 
\begin{align}
C^{baseline} &= \sum_{k=1}^K C_k^{testing} + C_{k^*}^{post} \label{baselinsetupstart} \\
with \ 
C_k^{testing} &\sim Binomial(\frac{M^{test}}{K}, CTR_k^{testing})  \label{baselinesetuptesting}\\
C_{k^*}^{post} &\sim Binomial(M^{post}, CTR_{k^*}^{testing})  \label{baselinsetupextrapolate}\\
k^* &= \argmax_{k\in\{1,...,K\}} C_k^{testing} \label{baselinsetupend}
\end{align} where $M^{test}$ and $M^{post}$ are the article impressions in the testing and post-testing period, respectively. $CTR_k^{testing}$ is the CTR for arm $k$ in the testing period. Expressions \eqref{baselinesetuptesting} and \eqref{baselinsetupextrapolate} mean $C_k^{testing}$ and $C_{k^*}^{post}$ are simulated from the corresponding Binomial distributions.

We set up this test-rollout baseline instead of directly using the empirical click data, in order to achieve a fair comparison with the bTS method. Suppose an article has two arms, with their \textit{testing period} empirical CTRs denoted as $CTR^{testing}_1$ and $CTR^{testing}_2$, and $CTR^{testing}_2 > CTR^{testing}_1$. Thus, arm $2$ is displayed to all traffic in the post-testing period, and we can calculate its ``overall'' empirical CTR, denoted as ${CTR^{overall}_2}$, by its empirical clicks divided by its empirical impressions during the overall article life. Note that ${CTR^{overall}_2}$ is not necessarily equal to $CTR^{testing}_2$ as illustrated in Section~\ref{limit2-fluctuateCTR}. 

The simulation for bTS uses $CTR^{testing}_1$ and $CTR^{testing}_2$ as the estimated success probability of corresponding arms as explained in Section~\ref{simulationpractice}, and it is not comparable with the empirical click data, where we can consider $CTR^{testing}_1$ and $CTR^{overall}_2$ as the underlying success probabilities. The possible gap between ${CTR^{testing}_2}$ and ${CTR^{overall}_2}$ can also lead to different click counts between bTS and empirical click data, even if bTS performs the same as current practice. On the other hand, the test-rollout baseline we set up in \eqref{baselinsetupstart} - \eqref{baselinsetupend} uses $CTR^{testing}_1$ and $CTR^{testing}_2$ as the underlying success probabilities, and thus is comparable with bTS.

The gain in clicks for bTS against the test-rollout baseline is summarized in Table~\ref{table:clickgain}. The gain in clicks is split into the ``first-hour'' period and the ``remaining-hour'' period, which correspond to the testing and post-testing periods in the test-rollout strategy. On average, bTS gives a 13.54\% increase in total clicks during the first hour. It also has competitive accuracy in selecting the correct optimal arm to exploit, indicated by 1\% increase in clicks during the post-testing period than test-rollout baseline. 

It is worth noting that this baseline is conservative - the actual gain after implementation is likely to be larger. The baseline setup assumes the CTRs of arms are consistent between testing and post-testing period. 
This assumption leads to dilution in click gain during the post-testing period: the baseline almost always displays the true optimal arm in the post-testing period due to the long period of exploration\footnote{There is a 97\% accuracy rate for the baseline to display the optimal arm during the post-testing period.}, and bTS cannot beat the baseline when the optimal arm is displayed in the baseline during the post-testing period. Thus, the 13.54\% click gain harvested in testing period is diluted to 3.69\% overall click gain. In practice, arm performance may change over time as mentioned in Section~\ref{limit2-fluctuateCTR}. The current test-rollout practice is not able to capture the fluctuation, while the bTS algorithm would detect the change and adjust the traffic accordingly, which would potentially gain more clicks. 

\begin{table}[!t]
\renewcommand{\arraystretch}{1.3}
\caption{Gain in clicks: bTS versus Test-rollout Baseline}
\label{table:clickgain}
\centering

	\begin{tabular}{| c | c | c | c |}
	\hline
     Total clicks & First hour & Remaining hours & Total \\ 
    \hline
    \hline
    \makecell{\% Increase in Clicks: \\ bTS versus \\ Test-rollout Baseline} & 13.54\% & 1.00\% & 3.69\%  \\
    \hline
	\end{tabular}
\end{table}

\subsection{Fewer Impressions on Sub-optimal Headlines}
\label{sec:fewer}
Another benefit of bTS is that it improves user experience by exposing much fewer sub-optimal headlines. We analyze the overall percentage decrease in impressions on sub-optimal headlines of the bTS method versus the test-rollout practice. For the test-rollout strategy which we use as the baseline, we assume that no traffic is allocated to sub-optimal headlines during the post-testing period. Even in this best-case scenario for the baseline, the sub-optimal headline impressions of the bTS method decrease by 71.53\% compared with the test-rollout practice.

%% file: 5.futurework.tex
In this paper, we propose to apply an MAB approach to headline testing where data are processed in batches due to their high volume and high velocity. Although the bTS algorithm is developed under the news article headline testing scenario, the parameters used to tune our model can be generalized to other real-world MAB problems, including marketing campaigns, one-time event optimizations, and user purchase/registration funnels. 

The stationary assumption beyond the first hour should be assessed after we have data on all arms beyond the first hour, which will be available after the bTS method is implemented. The current assumption is that $\theta_k$, the success probability for arm $k$, is constant over time. It is possible for $\theta_k$ to change over time, and this formulates a non-stationary bandit problem. Note that for non-stationary environments, the proposed bTS method is already an improvement of the test-rollout strategy, because it continuously tests headline variants throughout the active lifespan of news articles. There is room for further enhancement of the methodology to achieve better performance, especially when $\theta_k$ changes so significantly and rapidly that the order among $\theta_k$ for $k \in \{1,...,K\}$ flips multiple times during the article lifespan.

Clickbait detection and prevention are known challenges to media sites~\cite{chakraborty2016clickbait}. We are also motivated to explore more sophisticated metrics than raw clicks to be used as the reward of MAB. As an example, we may discount the clicks with a short dwell time, and the reward of each headline variation becomes a number between 0 and 1. We can then optimize this ``click-dwell'' reward using an adaptation of Bernoulli Thompson Sampling algorithm introduced in~\cite{agrawal2012analysis}, which supports the reward to be distributed in the interval [0,1]. When the discounting strategy of dwell time 
is properly selected, this should be helpful to mitigate the clickbait problem. 

At Yahoo Front Page, the top module hosts multiple articles. When headline testing is conducted simultaneously on multiple articles within this module, they may interact with each other. That is, the success probability $\theta$ of a headline variant may be influenced by the titles of the remaining articles in the same module. This motivates us to extend our research and develop an MAB testing framework that optimizes the article titles for the entire module, \textit{e.g.,} with a multivariate MAB algorithm introduced in~\cite{hill2017amazon}.

%% file: 6.acknowledgment.tex
The authors would like to express their gratitude to Russell Chen, Shriram Kumar, Jigar Patel, and Rupert Wu for valuable discussions; to Virendra Pratap Singh for his help with empirical data collection; and to Sameer Raheja and Kelly Hirano for their support of this study.